\documentclass[10pt,twocolumn,letterpaper]{article}

\usepackage{iccv}              

\usepackage{tablefootnote}
\usepackage{algorithm}
\usepackage{algorithmic}
\usepackage{url}            
\usepackage{booktabs}       
\usepackage{amsfonts}       
\usepackage{nicefrac}       
\usepackage{microtype}      
\usepackage{xcolor}         

\usepackage{amsmath,amsfonts,bm}

\def\eqref#1{equation~\ref{#1}}

\def\1{\bm{1}}

\def\rvc{{\mathbf{c}}}

\def\rvp{{\mathbf{p}}}

\def\rvt{{\mathbf{t}}}

\def\ermT{{\textnormal{T}}}

\DeclareMathAlphabet{\mathsfit}{\encodingdefault}{\sfdefault}{m}{sl}
\SetMathAlphabet{\mathsfit}{bold}{\encodingdefault}{\sfdefault}{bx}{n}

\def\gC{{\mathcal{C}}}
\def\gD{{\mathcal{D}}}

\def\gL{{\mathcal{L}}}

\def\gW{{\mathcal{W}}}

\def\sR{{\mathbb{R}}}

\newcommand*{\myDots}{\ifmmode\mathellipsis\else.\kern-0.13em.\kern-0.13em.\fi} 

\usepackage{graphicx}
\usepackage[draft]{pgf}
\usepackage{dsfont}
\usepackage{subcaption}
\usepackage{multirow}
\usepackage{soul}
\usepackage{comment}
\usepackage{multirow}
\usepackage{algorithm}
\usepackage{algorithmic}
\usepackage{wrapfig}
\usepackage{verbatim}
\usepackage{caption}
\usepackage{bbm}

\definecolor{iccvblue}{rgb}{0.21,0.49,0.74}
\usepackage[pagebackref,breaklinks,colorlinks,allcolors=iccvblue]{hyperref}

\def\paperID{xxx} 
\def\confName{ICCV}
\def\confYear{2025}

\title{Online Test-time Adaptation for 3D Human Pose Estimation: \\ A Practical Perspective with Estimated 2D Poses}

\author{%
Qiuxia Lin$^1$ \hspace{1cm}
Kerui Gu$^1$ \hspace{1cm}
Linlin Yang$^{2,3}$ \hspace{1cm}
Angela Yao$^1$\\
$^1$Department of Computer Science, National University of Singapore\\
$^2$State Key Laboratory of Media Convergence and Communication, CUC\\
$^3$School of Information and Communication Engineering, CUC\\
{\tt\small\{qiuxia, keruigu, ayao\}@comp.nus.edu.sg}
}

\begin{document}
\maketitle

\begin{abstract}
Online test-time adaptation for 3D human pose estimation is used for video streams that differ from training data. Ground truth 2D poses are used for adaptation, but only estimated 2D poses are available in practice. This paper addresses adapting models to streaming videos with estimated 2D poses. 
Comparing adaptations reveals the challenge of limiting estimation errors while preserving accurate pose information.
To this end, we propose adaptive aggregation, a two-stage optimization, 
and local augmentation for handling varying levels of estimated pose error. 
First, we perform adaptive aggregation across videos to initialize the model state with labeled representative samples. Within each video, we use a two-stage optimization to benefit from 2D fitting while minimizing the impact of erroneous updates. Second, we employ local augmentation, using adjacent confident samples to update the model before adapting to the current non-confident sample. 
Our method surpasses state-of-the-art by a large margin, advancing adaptation towards more practical settings of using estimated 2D poses.
\end{abstract}    
\section{Introduction}

Human pose estimation is common in applications where data arrives in video streams. 
Directly applying pre-trained models can be challenging if the video streams differ from the training data.
Online test-time adaptation (OTTA) offers a practical solution 
because it continuously update the model based on the 
new data 
from the testing phase~\cite{tent, liu2023vida, wang2022continual}. 
In 3D pose estimation, a typical strategy is to leverage 2D poses as a weak supervision, assuming that the poses are accurate~\cite{openpose, hrnet}. 
Previous works~\cite{boa,dynaboa} show an idealized proof-of-concept with ground truth poses.
Using estimated poses, as the task is intended, however, 
is more challenging; applied naively, it leads to catastrophic failures where errors even increase (see Table~\ref{tab:study} (a)).  This work investigates the underlying reasons and tackles the challenges of adapting 3D pose models to streaming videos with estimated 2D poses.

The conventional setting uses the  adapted model from the previous frame as an initial model and then adapts for the current frame based on the current 2D poses.
~\cite{tonioni2019real,volpi2022road, dynaboa} 
{While feasible with ground truth 2D poses, this ``single stream'' pipeline breaks down with noisy, estimated poses. 
The breakdown is exacerbated by the limited number of frames/poses used in the update. 
{This presents the challenge of limiting the impact of 
errors while preserving pose information.}  
Ideally, the adaptation strategy should be adjusted depending on the 2D pose accuracy.
In practice, as the accuracy is unknown, we use the 2D estimator confidence as a proxy~\cite{eft,gu2023calibration} to partition the keypoints and propose dedicated adaptation solutions for confident and non-confident poses.

First, to limit the impact of estimated 2D pose errors, we propose an alternative ``per-video stream'' baseline. This baseline re-initializes the model state back to the pre-trained model and performs a separate streaming update on each video. Such an approach eliminates incorrect updates accumulated from previous videos~\cite{wang2022continual}. Yet a full re-initialization 
wastes the opportunity to learn from the adaptations of the streamed test domain. 
To concurrently suppress error accumulation while leveraging historical data,  we strengthen the ``per-video stream'' baseline with an adaptive aggregation.   The aggregation is performed before the adaptation of each video starts. 
Specifically, we select representative frames from historical data based on accuracy and diversity. 
We then leverage 2D poses from non-confident samples and both 2D and 3D poses from confident samples.

{For each video, adapting to 2D poses inevitably creates a trade-off: insufficient adaptation loses the opportunity to benefit from the 2D data, while too much can introduce error and over-fitting.}
To address this, we propose a two-stage optimization for each test sample. In the first stage, we temporarily stop the update based on feature discrepancies.  We copy and store the model as the initialization for the next sample to prevent 
2D errors from affecting future samples; in the second stage, the update continues until the projected and estimated 2D poses closely align, providing the 3D prediction for the current sample. Separating the adaptation reduces the impact of error while maintaining fitting benefits.

Secondly, we observe that pose estimation may fail catastrophically in scenes with occlusions, truncations, and rare poses. The failures are sporadic and interspersed throughout video sequences while past samples may be less problematic.
In this case, we propose local augmentation to mimic the challenging conditions while leveraging confident adjacent samples.  
Limiting to only nearby confident samples ensures that the adaptation process is efficient and directly relevant to the failed samples. 
Summarizing our contributions, we:
\begin{itemize}
    \item Reveal the key challenge that limits the naive use of estimated 2D poses for 3D human pose adaptation 
    on streaming videos -- limiting error propagation while preserving information from estimated 2D poses.
    
    \item Propose adaptive aggregation and two-stage optimization to limit the impact of errors while retaining accurate pose information. This involves re-initializing and enhancing the model with representative historical data and a separate adaptation process for each test sample.
    
    \item Propose local augmentation to leverage confident samples to improve performance on non-confident ones. 

    \item Achieve up to a 18.0\% MPJPE improvement over state-of-the-art, 
    highlighting the benefits of different handling strategies for confident and non-confident poses. 
\end{itemize}

\section{Related Work}
\textbf{3D Human Mesh Recovery with 2D Labels.}
Obtaining accurate 3D labels for in-the-wild data is challenging, so using 2D poses to refine 3D mesh recovery results is a common strategy.
The refinement is typically achieved by a projection loss that enforces 2D alignment between the 2D poses and projected 3D mesh~\cite{simplify, eft, li2022cliff, KITRO}.
SMPLify~\cite{simplify} incorporates additional constraints to optimize the SMPL parameters for fitting 2D keypoints. EFT~\cite{eft} instead refines these predictions by fine-tuning the HMR estimation network weights.
While effective, these approaches are typically frame-based and do not consider the streaming nature of video data. 

Other approaches like DynaBOA~\cite{dynaboa} and ISO~\cite{iso} perform streaming adaptation. However, they are not robust because they require 2D ground truth poses.  Their performance suffers with estimated poses and this is the setting that we target.   
{Our work, like DynaBOA, leverage guidance by storing samples. However, a key difference is that we use only unlabeled target samples and store a limited number of representatives, which we feel is better aligned with the spirit of OTTA. In contrast, DynaBOA stores {labeled training samples from the source domain}. 
}

\noindent \textbf{Unsupervised Adaptation on Streaming Data.}
This work focuses on video -- a common form of streaming data where consecutive frames share a high similarity.
It is natural to consider sequential adaptation, which has been used in other tasks such as object tracking~\cite{park2018meta}, semantic segmentation~\cite{wang2022continual,liu2023vida}, and object recognition~\cite{lin2023video}.
In these tasks, only the streamed video is accessible, so the main challenge is to avoid over-fitting while still learning (unsupervised) from the video stream.  Commonly used strategies include feature alignment~\cite{lin2023video}, meta-learning~\cite{park2018meta}, partial fine-tuning~\cite{liu2023vida}, and resetting~\cite{wang2022continual}.
Adapting 3D HMR in streaming video differs slightly because 2D detected poses are easily accessible and can be used as weak forms of supervision for the adaptation. 
Different from~\cite{boa, dynaboa} that uses 2D poses indiscriminately (since they only consider ground truth 2D poses), we target noisy 2D to effectively exploit the beneficial information while mitigating the impact of noisy data.

\noindent \textbf{Learning with Noisy Labels.} 
A straightforward approach is to identify noisy samples and either exclude~\cite{han2018co} or downweight them~\cite{liu2015reweight} during model training.
Some works consider model output to correct the 
loss dynamically~\cite{arazo2019soft,tanaka2018hard}.
Alternatively, noisy labels can be considered by improving model robustness, \eg by augmenting the clean labels~\cite{hoang2024improving}. ~\cite{cycleadapt} corrects 3D pose predictions generated from noisy 2D labels by cyclically performing motion denoising with the pose estimation. 
Given the unpredictability of noise and constraints on training time, our method tries to filter out noisy samples while leveraging the similarity between consecutive frame samples to improve the model robustness to noise.
\section{OTTA on 3D Human Pose Estimation}

\begin{table*}[t]
\flushleft
\begin{minipage}[c]{0.4\linewidth}
    \small
    \scalebox{0.94}{
    \begin{tabular}{l|l|c|c}
    \toprule
    Method & 2D pose & MPJPE  & PA-MPJPE  \\
    \midrule
    No adaptation & - & 233.3 & 116.8 \\
    \midrule
    Single stream & GT  & 65.5 & 40.4 \\
    Per-video stream & GT & 78.6 & 46.3 \\
    \midrule
    Single stream & Estimated &  138.5	& 79.2 \\
    Per-video stream & Estimated & 120.4	& 73.8\\
    \bottomrule
    \end{tabular}%
    }%
    \end{minipage}
    \begin{minipage}[c]{0.28\linewidth}
    \centering
        \scalebox{0.98}{
        \includegraphics[width=\linewidth]{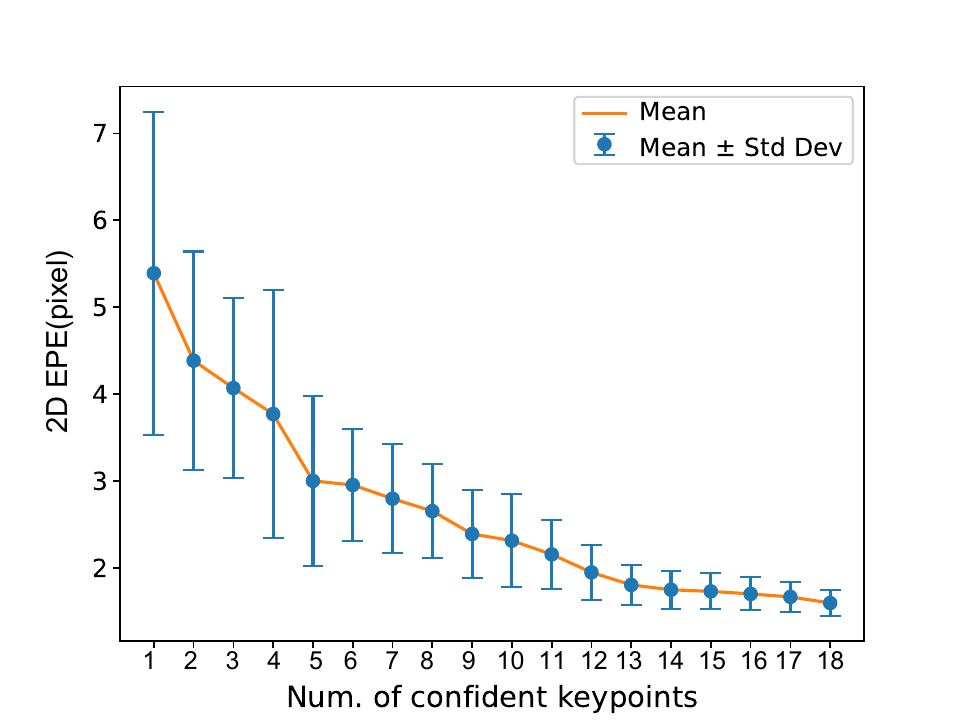}
        }
    \end{minipage}
    \begin{minipage}[c]{0.28\linewidth}
    \centering
        \scalebox{0.98}{
        \includegraphics[width=\linewidth]{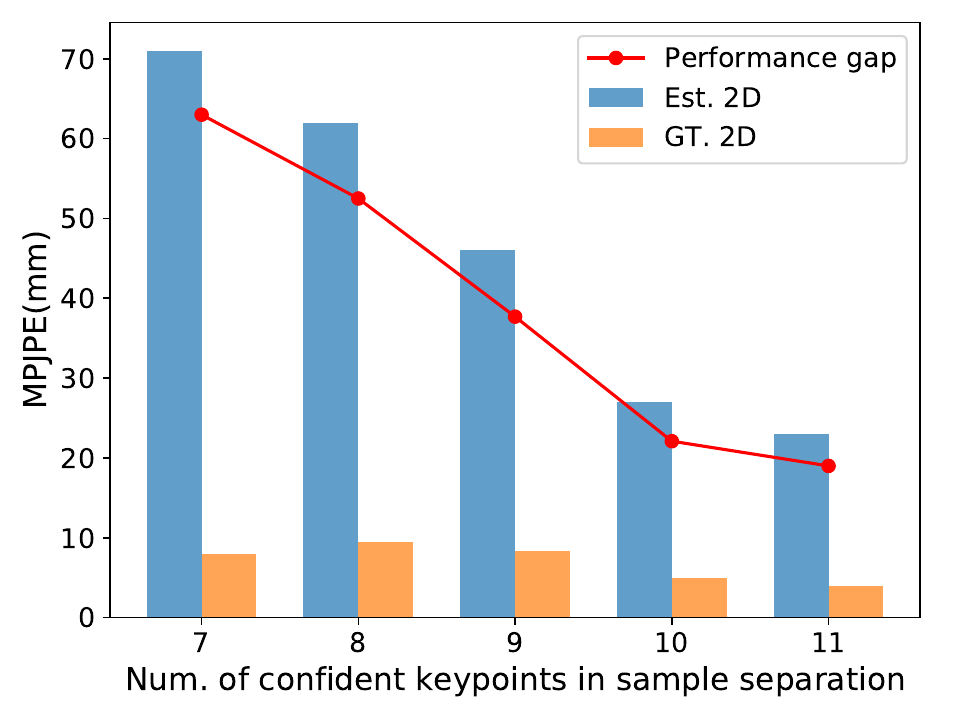}
        }
    \end{minipage}
    \begin{minipage}[t]{0.4\linewidth} 
    \centering {\small (a) Comparison of adaptation settings.}%
    \end{minipage}%
    \begin{minipage}[t]{0.3\linewidth} 
    \centering {\small (b) Keypoint confidence vs. pose error.} \\
    \end{minipage}%
    \begin{minipage}[t]{0.3\linewidth} 
    \centering {\small (c) {Estimated vs. GT Performance}}
   
    \end{minipage}%
    
\caption{(a) Adaption of DynBOA~\cite{dynaboa} on the 3DPW~\cite{3dpw} test set using ``single stream'' vs ``per-video stream'' pipelines, with ground truth 2D and poses estimated with OpenPose~\cite{openpose}. 
Adaptation outperforms no adaptation; with ground truth poses, the single stream is better than the per-video stream, but the trend reverses for noisy, estimated poses due to error accumulation.
(b) Relationship between the number of confident keypoints and the 2D EPE (pixel) on 3DPW test set. 
High 2D error samples can be filtered out by keypoint confidence. 
(c) A significantly larger performance gap (MPJPE [mm]) between using estimated 2D and ground truth 2D labels is {observed in samples with fewer confident keypoints}.
Results are based on the 3DPW test set with ``single stream'' pipeline. 
}
\label{tab:study}%
\vspace{-0.3cm}
\end{table*}

\subsection{Preliminaries}
\textbf{Framework.}
We construct a CNN with an iterative regression module as the 3D human pose and shape estimation model $\mathcal{M}_{\phi}$, as per~\cite{hmrnet}. The model takes an RGB {image} as input and gives
$\{\boldsymbol{\hat{\theta}}, \boldsymbol{\hat{\beta}}, \hat{\rvt}\}$ as output, where $\boldsymbol{\hat{\theta}}\in\sR^{72}, \boldsymbol{\hat{\beta}}\in \sR^{10}$ 
are SMPL model~\cite{smpl} pose and shape parameters, and $\hat{\rvt}\in \sR^{3}$ represents the root joint's camera-relative translation.

\noindent \textbf{2D weak supervision.} The 3D human pose is refined by projection into 2D and comparison with 2D pose labels. 
Based on the recovered mesh from $\{\boldsymbol{\hat{\theta}}, \boldsymbol{\hat{\beta}}\}$, forward kinematics can be performed to obtain the estimated 3D pose $\hat{\rvp}_\mathrm{3d}\in \mathbb{R}^{J\times 3}$ where $J$ is the number of keypoints.  The 3D pose $\hat{\rvp}_\mathrm{3d}$ is projected into 2D pose $\hat{\rvp}_\mathrm{2d}\in\mathbb{R}^{J\times 2}$ based on the camera projection function $\Pi$ and the camera translation $\hat{\rvt}$, before comparing with the ground truth $\rvp_\mathrm{2d}$ with a projection loss:
\begin{equation}
\begin{aligned}
\gL_{2D}=||\Pi({\hat{{\rvp}}_\mathrm{3d},{\hat{\rvt}}}),\rvp_\mathrm{2d}||_2^2.
\end{aligned}
\label{Eq:2D}
\end{equation}
Above, the ground truth pose 
can also be replaced by an estimated pose $\Tilde{\rvp}_\mathrm{2d}\in\sR^{J\times2}$ from an off-the-shelf 2D pose estimation framework~\cite{openpose, hrnet, xu2022vitpose}.  Typically, such frameworks also provide a confidence measure $\rvc\in\sR^{J}$ that roughly represents the prediction accuracy.

\noindent \textbf{Problem Formulation.} 
Consider a model $\mathcal{M}_{\phi_0}$ pre-trained on a labeled 3D human dataset $\gD_s$ with ground truth poses $\{\boldsymbol{\theta}, \boldsymbol{\beta}\}$. 
Online test-time adaptation is performed on the test data $\mathcal{D}_t = \{V_1, V_2, \ldots, V_n\}$ where the videos can differ in backgrounds, camera views, etc., in real scenarios. The sequence $V_i = \{x_{i1}, x_{i2}, \ldots, x_{im}\}$ denotes the frames in video $i$, while 
\(x_{ij}\) denotes frame $j$ from video \(V_i\). The test frames arrive sequentially and for each frame \(x_{ij}\), the model is adapted from \(\mathcal{M}_{\phi_{i(j-1)}}\) to \(\mathcal{M}_{\phi_{ij}}\), with the help of only previous frames \(\{x_{i1}, \ldots, x_{i(j-1)}, x_{ij}\}\) and their 2D poses. Previous work adopted a ``single stream'' pipeline that $\mathcal{M}_{\phi}$ is updated continuously by the incoming frames with 2D ground truth poses, not only within each video but also across videos. However, in practice, 2D ground truth poses for arbitrary videos are unavailable. 
Our work tackles adaptation with estimated 2D poses in a realistic setting.

\subsection{Challenges of using Estimated 2D Poses}

What happens when ground truth 2D poses are replaced with estimated ones for adapting the 3D model?  The following experiments on the state-of-the-art DynaBOA~\cite{dynaboa} highlight some key differences that motivate our  development.
    
Table.~\ref{tab:study}(a) shows that using estimated 2D poses in a ``single stream'' pipeline approximately doubles the errors compared to using ground truth poses. This pipeline, which feeds in videos consecutively, is common practice in OTTA scenarios~\cite{tonioni2019learning, volpi2022road, dynaboa}, regardless of differences across the videos.  
We consider an alternate ``per-video stream"\footnote{Our naming convention aligns to the use of established datasets with individual videos; in real-world application streams, natural breaks also occur, \eg switching cameras on/off, new sessions, new users, etc.} baseline, which 
resets the model to $\mathcal{M}_{\phi_0}$ when video $V_i$ ends.
Such a simple change (per-video stream w/ Est. 2D) significantly improves adaptation results over the ``single stream'' with estimated 2D poses.

A key difference between ground truth and estimated 2D poses arises in challenging scenarios like occlusions, truncations, or rare poses.  In these cases, pose estimation often fails; it also leads to keypoints with low confidence estimates. 
Poses with more confident keypoints (we use a confidence threshold of 0.8) generally have lower 2D errors, as shown in Table.~\ref{tab:study} (b). We consider test poses as confident if they have more than 10 keypoints with confidence estimates above 0.8.
Table~\ref{tab:study} (c) shows an increasing performance gap of 3D adaptation error between using estimated 2D and ground truth 2D as the number of confident keypoints decreases.
It highlights the need to improve these more challenging samples.

\section{Methodology}

\begin{figure*}[!h]
    \centering
    \includegraphics[width=0.9\textwidth]{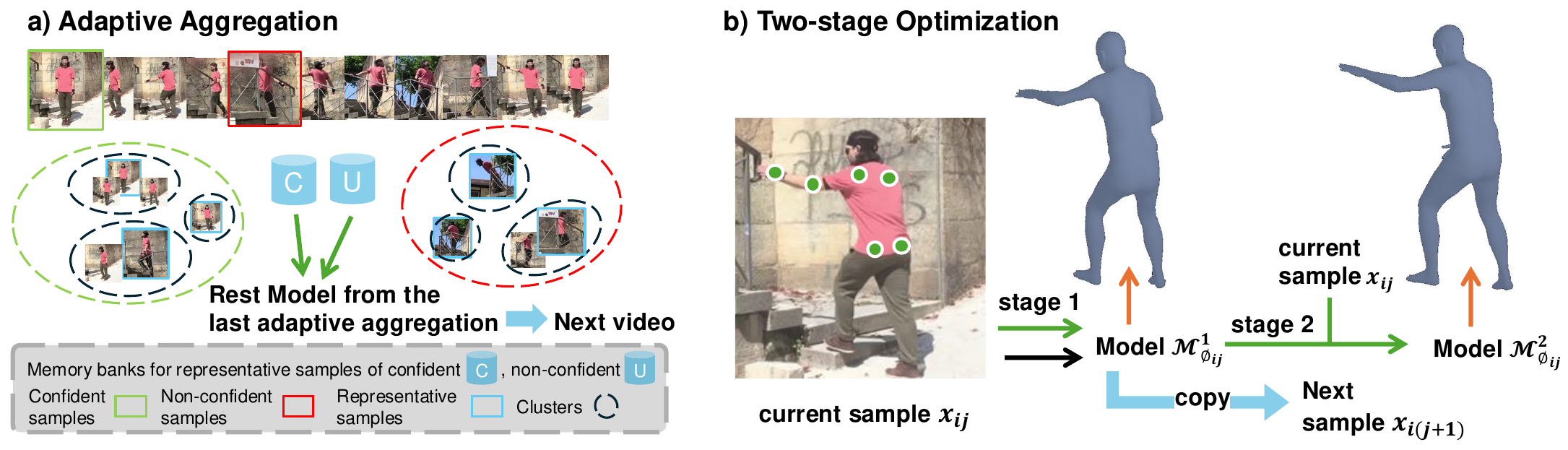}

    \caption{
    (a) Adaptive Aggregation: Representative samples from both confident and non-confident data are selected using spherical K-means, with memory banks storing these samples for limited retrieval during video transitions. 
    (b) Two-stage Optimization: Each sample undergoes two-stage adaptation to maintain 2D fitting and minimize incorrect updates.
   }
    \label{fig:architecture_ab}
\end{figure*}

We propose a unified framework to reduce the impact of noisy 2D estimations and improve the 3D adaptation performance on non-confident 2D estimates. First, we introduce an  \textbf{Adaptive Aggregation} (Sec.~\ref{sec:adaptive_aggregation}) that changes the streaming pipeline to update the model only with reliable and representative samples. Secondly, we introduce a \textbf{Two-Stage Optimization} process (Sec.~\ref{sec:two_stage}) during the adaptation of each sample. Third, to improve performance on non-confident samples, we propose \textbf{Local Augmentation} (Sec.~\ref{sec:local_aug}) by leveraging nearby confident samples.

\subsection{Adaptive Aggregation}\label{sec:adaptive_aggregation}

Continuous adaptations over long video streams makes the model more susceptible to inaccurate 2D estimates.
To address this, we first base our method on the ``per-video stream", which resets the model each time a new video is encountered.
To leverage historical data from other videos, however, we {incrementally update} the reset model using representative samples selected from past videos.

The simplest way to incorporate past data is through uniform sampling. However, confident and non-confident samples are not uniformly distributed and present different properties. 
Confident samples have smaller pose errors, but exhibit limited poses and scenes.  Conversely non-confident samples have 
higher pose errors but are draw from a richer set of poses and scenes (occlusions, truncation). 
To use both types of samples, we propose three key components in Adaptive Aggregation after resetting the model: 1) Balanced sample selection, 2) Pose clustering, and 3) an Adaptive pseudo-labeling loss.

\noindent \textbf{Balanced sample selection.} For the reset model to learn the distribution from both confident and non-confident samples, we split the video data into these two subsets according to the number of confident estimated 2D keypoints and select a balanced number of samples from each subset. To further distinguish the selection,  
we assign a sampling weight for each sample based on the average 2D confidence: 
\begin{equation}
    \gW= \sum_{j=1}^{J}{\rvc_j}/J.
    ~\label{Eq:sample_weight}
\end{equation}

\noindent \textbf{Pose clustering.} Within both the confident and non-confident subsets, the high sampling weights are associated with a limited set of poses. As such, we apply Spherical K-means clustering~\cite{hornik2012spherical} on the 3D poses in each subset.
As each sample is assigned a sampling weight (See Eq.~\ref{Eq:sample_weight}), we order the samples within each cluster by weight and select a specific number based on the ratio of the cluster size to the intended total sampling size $\gC_iN_v/\sum_{i=1}^{N_\gC}\gC_i$, where $\gC_i$ represents the size of cluster $i$, $N_\gC$ is the number of clusters and $N_v$ is the required total sampling number in each video. 
The selected $N_v$ samples are stored within the subset $\gD_m$ for use in the adaptive aggregation.

\noindent \textbf{Adaptive pseudo-labeling loss.} Since there is no ground truth label and the adapted results may be supervised by incorrect 2D poses, we adopt an adaptive pseudo-labeling loss.  Specifically, we apply a strong pseudo-labeling loss based on 2D and 3D labels for the confident samples and a weak pseudo-labeling loss with only 2D labels for the non-confident samples:
\begin{equation}
\begin{aligned}
\gL_{adapt}&=||\ermT_w^{-1}(\hat{{\rvp}}_\mathrm{2d}'),\Tilde{\rvp}_\mathrm{2d}||_2^2+\mathbbm{1}\lambda_2||\ermT_w^{-1}(\hat{{\rvp}}_\mathrm{3d}'), \hat{{\rvp}}_\mathrm{3d}||_2^2\\
&+\lambda_1(E_{\boldsymbol{\hat{\theta}}}({\boldsymbol{\hat{\theta}}})+E_{\boldsymbol{\hat{\beta}}}({\boldsymbol{\hat{\beta}}})).
\end{aligned}
\label{Eq:adaptive_pl}
\end{equation}

\begin{figure*}[!h]
    \centering
    \includegraphics[width=0.9\textwidth]{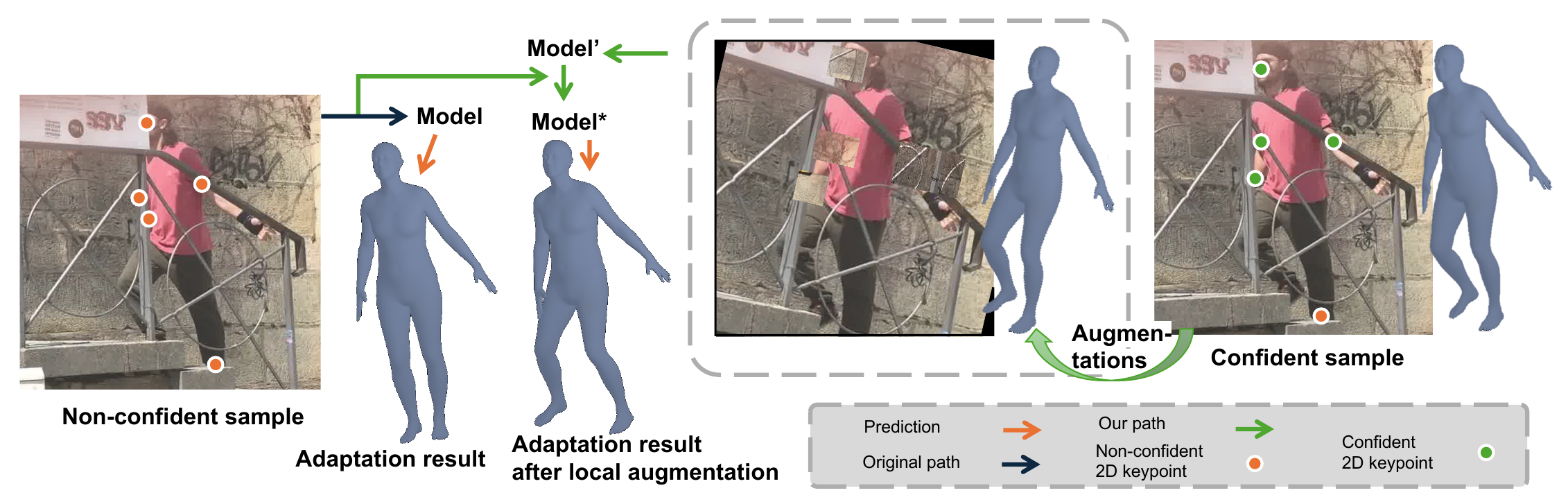}

    \caption{
    Local augmentation.
    We augment the temporally adjacent confident samples to simulate the hard features in non-confident samples.
    The transformed predictions are used to guide the model before adapting to the non-confident sample.
    Specifically, we first adapt the model using augmented confident samples to obtain Model'. Then, Model' is further adapted with non-confident samples to obtain the final Model*, which is used for prediction.
    }
    \label{fig:architecture_c}
    \vspace{-0.2in}
\end{figure*}

\noindent Above, the first and the second loss term enforces 2D and 3D consistency, respectively; the third term is a prior 
applied to the shape and pose parameters, similar to ~\cite{spin, boa}. The indicator $\mathbbm{1}$ indicates the additional strong pseudo-labeling term for the confident samples. 
In both terms, we encourage consistency between the predicted 
$\hat{{\rvp}}_\mathrm{3d}$, estimated $\Tilde{\rvp}_\mathrm{2d}$ of the original image and transformed-back predictions $\hat{{\rvp}}_\mathrm{3d}'$ , $\hat{{\rvp}}_\mathrm{2d}'$ of the augmented image. $\ermT_w$ represents the geometric transformation including rotation, scaling, translation, that we apply on the original images to get corresponding augmented ones. $\ermT_w^{-1}$ denotes the reversed transformation.
Symbols with a prime superscript denote the predictions on the augmented images, to distinguish them from predictions on the original image.
For non-confident samples, we apply weak pseudo-labelling loss for only utilizing 2D information.

We perform adaptive aggregation when the video changes, selecting  $N_v$ samples from $\gD_m$ to train the reset model using adaptive pseudo-labeling loss. To prevent earlier samples added to $\gD_m$ from being selected more frequently than others, we apply a selection probability on each sample in $\gD_m$, which is inversely proportional to the number of times it has been previously chosen. Instead of following ``single stream'' to update model across videos, which transits the incorrect updates from the previous video to the next video, we use adapted results that are representative 
to incrementally update the model across videos.

\subsection{Two-stage Optimization}\label{sec:two_stage}
Off-the-shelf 2D pose estimators can often provide more accurate 2D estimates than those projected from 3D human pose models, especially when there is a domain gap~\cite{3dhp}. This information can be valuable for refining 3D poses. However, errors in estimated 2D poses are inevitable. Adapting the model using these wrong 2D poses can lead to harmful updates, specifically causing low generalization ability to subsequent samples. 
To address this, we propose a two-stage optimization strategy to enhance the fitting only on the current sample and mitigate the negative influence on the following samples simultaneously. 

Given the estimated 2D poses, we first update the model parameters based on the 2D projections of the estimated 3D pose, along with the pose and shape priors.
\begin{equation}
\begin{aligned}
\gL_{proj}=||\Pi({\hat{{\rvp}}_\mathrm{3d},{\hat{\rvt}}}),\Tilde{\rvp}_\mathrm{2d}||_2^2+\lambda_1(E_{\boldsymbol{\hat{\theta}}}({\boldsymbol{\hat{\theta}}})+E_{\boldsymbol{\hat{\beta}}}({\boldsymbol{\hat{\beta}}}))
\end{aligned}
\label{Eq:2d_project}
\end{equation}
We aim for the projected 2D poses (from the estimated 3D poses) to align with the estimated 2D poses. After several 
alignments, due to the potential errors in the estimated 2D poses, the model may 
start to fail on subsequent samples. As such, we consider a first stage with early stopping~\cite{early-stopping} when the feature changes stabilize~\cite{dynaboa}. 
We measure the cosine similarity between features in a specific layer before and after each iteration based on Eq.~\ref{Eq:2d_project} with current sample and stop the optimization when the similarity falls below a certain threshold.
As illustrated in Fig.~\ref{fig:architecture_ab}(b), as the first stage finishes for the input frame $x_{ij}$, we have the adapted model $\mathcal{M}_{\phi_{ij}}^1$ and pass it to the next sample $x_{i(j+1)}$. To achieve better aligned 2D predictions, we then continue updating the model to $\mathcal{M}_{\phi_{ij}}^2$
until the 2D fitting loss is minimized to the required value for better-aligned 2D predictions. The final 3D adaptation results of $x_{ij}$ is obtained by $\mathcal{M}_{\phi_{ij}}^2$ which will not be utilized in further samples.

\subsection{Local Augmentation}\label{sec:local_aug}

For the case where the 2D estimator fails, we designed a local augmentation strategy to leverage confident samples to assist the model prediction on the non-confident samples. The rationale is that 
the non-confident samples tend to appear unpredictably, often during brief occlusions or when the human moves out of the camera's view, while surrounding samples remain unaffected by these challenges, which results in sporadic distribution of non-confident samples.

How can we leverage the previous confident sample with relatively accurate 2D and 3D labels to help the prediction of the current non-confident sample? 
We observe that the non-confident samples often feature occlusions and truncations, so we apply strong augmentations on previous confident samples to improve model robustness. 
We provide an illustration in Fig.~\ref{fig:architecture_c}.
Strong augmentations, denoted by $\ermT_s$, includes geometry and texture augmentation, which are {rotation, scaling, translation, color normalization}, image truncation, and keypoint occlusion. Considering the pose similarity, we only augment the confident sample within a short temporal window before the non-confident sample.

Preliminary experiments show that a window of five frames is appropriate (See Supp.).
If a specific keypoint is confident in the former but non-confident in the latter, we will occlude it in the former to simulate the challenges in the latter.

Suppose the model 3D prediction on the original unaugmented confident sample is $\hat{{\rvp}}_\mathrm{3d}$ and the estimated 2D is $\Tilde{\rvp}_\mathrm{2d}$, the training can thus be supervised by enforcing the consistency between the original predictions and transform-backed predictions $\ermT_s^{-1}(\hat{{\rvp}}_\mathrm{2d}')$, $\ermT_s^{-1}(\hat{{\rvp}}_\mathrm{3d}')$ on the augmented image.

\begin{equation}
\begin{aligned}
\gL_{aug}=||\ermT_s^{-1}(\hat{{\rvp}}_\mathrm{2d}'),\Tilde{\rvp}_\mathrm{2d}||_2^2+\lambda_2||\ermT_s^{-1}(\hat{{\rvp}}_\mathrm{3d}'), \hat{{\rvp}}_\mathrm{3d}||_2^2.
\end{aligned}
\label{Eq:local_aug}
\end{equation}

\begin{table*}[!h]
  \centering
  
\setlength{\tabcolsep}{10pt}
    \scalebox{0.9}{
    \begin{tabular}{l|c|cc|cc}
    \toprule
     \multirow{2}{*}{2D Pose Estimator} & \multirow{2}{*}{Method} & \multicolumn{2}{c|}{\textsl{Human3.6M $\rightarrow$ 3DPW}} & \multicolumn{2}{c}{\textsl{Human3.6M $\rightarrow$ 3DHP}}  \\
     & & MPJPE & PA-MPJPE & MPJPE & PA-MPJPE \\
     \midrule
    \multirow{2}[2]{*}{OpenPose~\cite{openpose}} & DynaBOA  &  138.5	& 79.2  & 132.3	& 86.4 \\
              & CycleAdapt*  & 123.6 & 
 70.5 & 130.1 & 85.2 \\
          & Ours  & 101.3 & 64.5 & 122.4 & 82.6 \\
    \midrule
    \multirow{2}[2]{*}{HRNet~\cite{hrnet}} & DynaBOA & 108.0 & 63.9 & 130.3 & 86.8 \\
     & CycleAdapt*  & 102.8 & 61.2 & 122.8 & 83.5 \\
          & Ours  & 95.3 & 60.6 & 120.1 & 80.7 \\
    \midrule
    \multirow{2}[2]{*}{VitPose~\cite{xu2022vitpose}} & DynaBOA & 91.8 & 56.6 & 118.7 & 77.7 \\
     & CycleAdapt*  & 90.5 & 55.8 & 115.3 & 75.6 \\
          & Ours  & 87.2  & 53.5 & 112.1 & 73.2 \\
    \midrule
    \multirow{2}[2]{*}{Ground Truth} & DynaBOA & 65.5  & 40.4  & 101.5 & 66.1 \\
     & CycleAdapt*  & 67.1 & 44.3 & 102.1 & 67.5 \\
          & Ours  & 64.9  & 40.6  & 100.8 & 65.4 \\
    \bottomrule
    \end{tabular}%
}
    \caption{Robustness to different estimated 2D~\cite{openpose, hrnet, xu2022vitpose}. Our method demonstrates robustness across 2D inputs of varying accuracy.  *Modified CycleAdapt source code to accommodate the online setting.
    }    \label{tab:2d_robustness}
    \vspace{-0.15in}
\end{table*}%

\subsection{Method Overview} 

\textbf{Mean-teacher training.} 
We incorporate mean-teacher training~\cite{meanteacher} as it is {widely used in unsupervised learning}~\cite{synanimal21,yuan2023robust,dynaboa}. It can enhance stability and mitigate the impact of unreliable test data.
It involves maintaining both a student and a teacher model.
The student model is updated using designed losses, while the teacher model is updated by temporal weight ensembling of the student model.  The reported results come from the student model.

\noindent \textbf{Overall pipeline.} {Given several test videos $\mathcal{D}_t = \{V_1, \dots, V_n\}$, we adapt the model on them sequentially.
For each arriving sample $x_{ij}$ in video $i$, we apply specific strategies based on its confidence.
Confident sample is directly adapted with two-stage optimization (Eq.~\ref{Eq:2d_project}) from model state $\mathcal{M}_{\phi_{i(j-1)}}$ to obtain $\mathcal{M}_{\phi_{ij}}^2$ for inference and $\mathcal{M}_{\phi_{ij}}^1$ for the next sample $x_{i(j+1)}$.
If the current sample is non-confident, we use previously confident samples within the temporal window to update the model with local augmentation (Eq.~\ref{Eq:local_aug}) followed by the two-stage optimization.
Once adaptation for video $i$ is complete, representative samples are taken through balanced selection and clustering, and stored in $\gD_m$.
Prior to processing the next video $i+1$, the model is first reversed to the initial state of video $i$, \ie, $\mathcal{M}_{\phi_{i}}$.
We then enhance it with adaptive aggregation (Eq.~\ref{Eq:adaptive_pl}) through weighted sampling from $\gD_m$ to achieve initial state $\mathcal{M}_{\phi_{i+1}}$ for video $i+1$.
The overall illustration and algorithm can be found in Supplementary Materials.}
\section{Experiments}

\subsection{Setup, Datasets \& Evaluation}
\noindent \textbf{Setup:} We use various 2D estimators which provide 2D poses and keypoint confidences: OpenPose~\cite{openpose}, HRNet~\cite{hrnet}, VitPose~\cite{xu2022vitpose}. 
The backbone is HMR network~\cite{hmrnet}. 
Our method is implemented with Pytorch and trained using Adam optimizer with a momentum of 0.5 and an initial learning rate of $3\!\times\!10^{-6}$. The images are resized to $224\!\times\!224$.
The hyperparameters $\lambda_1, \lambda_2$ are set as 1e-4 and 10. The seed is 22. The cluster number $N_{\gC}\!=\!15$ and the required total sampling number $N_v\!=\!160$ in each video.
During adaptive aggregation, we will use Adam optimizer with a momentum of 0.7 and a learning rate of $3 \times 10^{-5}$. The batch size is 8 in adaptive aggregation and 1 in local augmentation and two-stage optimization.  
We compare with state-of-the-art test-time adaptation methods (DynaBOA~\cite{dynaboa}, ISO~\cite{iso}, DAPA~\cite{DAPA}, CycleAdapt~\cite{cycleadapt}) and optimization-based methods (SMPLify~\cite{simplify} and EFT~\cite{eft}). {Note that CycleAdapt is designed for (offline) test-time adaptation; they do not release source code for the online setting, and we modify their code to support it.}

\begin{table}[!t]
\scalebox{0.9}{
\begin{tabular}{l|cc|cc}
\toprule
 \multirow{2}{*}{Method} & \multicolumn{2}{c|}{\textsl{Human3.6M $\rightarrow$ 3DPW}} & \multicolumn{2}{c}{\textsl{Human3.6M $\rightarrow$ 3DHP}}  \\
 & MPJPE & PA-MPJPE & MPJPE & PA-MPJPE \\
 \midrule
 No adaptation & 233.3 & 116.8 & 218.4 & 118.4 \\
 \midrule
  DynaBOA~\cite{dynaboa} &  138.5	& 79.2  & 132.3	& 86.4
   \\
   CycleAdapt*~\cite{cycleadapt} 
    & 123.6 & 70.5 & 130.1 & 85.2 \\ 
\textcolor{gray}{    CycleAdapt}\tablefootnote{Reported from~\cite{cycleadapt}'s Supplementary Table B(a); results not reproducible from provided code.}\cite{cycleadapt} & \textcolor{gray}{90.3} & \textcolor{gray}{55.2} & - & - \\ 
  ISO~\cite{iso} & - &70.8 & -& - \\
  SMPLify~\cite{simplify} & 190.3	& 97.5	& 174.8 & 105.3 \\
  EFT~\cite{eft} & 143.6	& 83.2	& 152.3	& 92.1 \\
  Ours & \textbf{101.3} & \textbf{64.5} & \textbf{122.4} & \textbf{82.6} 			\\
\bottomrule
\end{tabular}}
\caption{For online TTA with estimated 2D poses from OpenPose~\cite{openpose}, our method performs best. *Modified CycleAdapt source code to accommodate the online setting.  
}
\label{tab:online_results_flops}
\vspace{-0.2in}
\end{table}

\begin{table*}[!t]
\flushleft
    \begin{minipage}[c]{0.45\linewidth}
    \centering
    \scalebox{0.9}{
    \begin{tabular}{l|c|c|c}
    \toprule
      \multicolumn{1}{l|}{Method} & \multicolumn{1}{c|}{All} & \multicolumn{1}{c|}{Conf.} & \multicolumn{1}{c}{Non-conf.} \\
    \midrule
    Per-video stream & 120.4 & 117.8 & 160.5 \\
    \midrule
    ~~+adaptive agg. & 105.2 & 103.6 & 124.7 \\
    ~~+local aug. & 107.5 & 106.8 & 130.5 \\
    ~~+two-stage & 106.1 & 104.3 & 133.8 \\
    \bottomrule
    \end{tabular}}%
    \end{minipage}
\begin{minipage}[c]{0.49\linewidth}
    \centering
    \scalebox{0.9}{
    \begin{tabular}{l|c|c|c}
    \toprule
          Method & \multicolumn{1}{c|}{weak pl} & \multicolumn{1}{c|}{strong pl} & \multicolumn{1}{c}{adaptive pl} \\
    \midrule
    Uniform sampling & 114.5 & 116.2 & 114.3 \\
    \midrule
    Weight-based sampling & 111.6 & 115.2 & 111.5 \\
    ~~+Balanced selection & 109.5 & 114.3 & 107.6 \\
    ~~~+Pose Clustering & 107.7	& 108.2 & 105.2 \\
    \bottomrule
    \end{tabular}%
          }
    \end{minipage}
    \begin{minipage}[t]{0.5\linewidth} 
    \centering {\small (a) Component-wise Ablations of MPJPE (mm)}
    \end{minipage}%
    \begin{minipage}[t]{0.45\linewidth} 
    \centering {\small (b) Adaptive Aggregation Ablation}
    \end{minipage}%
\caption{(a) Ablation study on the 3DPW test set with MPJPE (mm). Adaptive aggregation (+adaptive agg.) excelled overall. Local augmentation (+local aug.) is better for non-confident samples, while two-stage optimization benefits confident samples more. (b) The adaptive aggregation increases sample diversity and quality.}
\label{tab:pipelines_n_ablation}%
\vspace{-0.2in}
\end{table*}

\noindent \textbf{Datasets}: \textbf{Human3.6M}~\cite{human36m} comprises 3.6 million indoor human pose samples with static backgrounds with ground truth 2D and 3D poses.
Following protocol 1, we pre-train our model using samples from subjects 1, 5, 6, 7, and 8.
\textbf{3DPW}~\cite{3dpw} is a challenging in-the-wild human pose dataset with extensive occlusions, varied lighting conditions, and truncations. We utilize only its test data, which consists of 24 videos with 35,515 samples. 
\textbf{MPI-INF-3DHP (3DHP)}~\cite{3dhp} is another benchmark from which we use the test set to perform model adaptation. The test set contains 6 videos with 2,929 samples across diverse backgrounds including green screen, in-studio, and in-the-wild settings. The poses 
are generally more complex than those in 3DPW.

\noindent \textbf{Evaluation}: 
The Human3.6M dataset is used as training data for model pre-training with available annotations. The test set of 3DPW and 3DHP are used as unlabeled test data for model adaptation, thus we can build two test-time adaptation tasks \textsl{Human3.6M $\rightarrow$ 3DPW} and \textsl{Human3.6M $\rightarrow$ 3DHP} to evaluate the proposed method. The applied evaluation metrics are the Mean Per Joint Position Error (MPJPE) between the ground truth and the predicted 3D pose, and Procrustes Aligned MPJPE (PA-MPJPE) with additional 3D alignment. Both are reported in millimeters.

\subsection{Results of Online Test-Time Adaptation}~\label{sec:results}
\vspace{-0.2in}

\noindent \textbf{Results on 3DPW \& 3DHP.} 
Table~\ref{tab:online_results_flops}
compares our method with state-of-the-art on two online test-time adaptation tasks. 
With estimated 2D data, the optimization-based methods~\cite{simplify, eft} perform poorly. Although slightly better, test-time adaptation methods~\cite{iso, dynaboa, cycleadapt} also struggle. 
These results highlight the challenge of using estimated 2D poses to adapt 3D models. 
Our method performs best by effectively utilizing estimated 2D data to extract valuable information from the test data.

\noindent \textbf{Robustness to Different 2D Pose Estimators.}
Table~\ref{tab:2d_robustness} compares our method with DynaBOA and CycleAdapt using different 2D pose estimators~\cite{openpose, hrnet, xu2022vitpose}, with differing levels of accuracy. 
{Our method significantly outperforms DynaBOA on the 3DPW and 3DHP datasets.
CycleAdapt improves upon DynaBOA, which targets estimated 2D poses, by reducing noisy 2D poses through motion denoising. 
However, it still struggles in the online setting due to limited sample optimization with noisy 2D poses.  This leads to overfitting during the cyclic update between modules. 
We propose more refined strategies to handle such extreme cases with noisy labels.}
The result verifies the ability of our method to improve 3D prediction even when the weak supervising labels (the 2D poses) has errors. As 2D accuracy improves, our method outperforms comparisons and stays competitive with ground truth 2D, showing its robustness.

\begin{figure}[!t]
\centering
\includegraphics[width=0.85\linewidth]{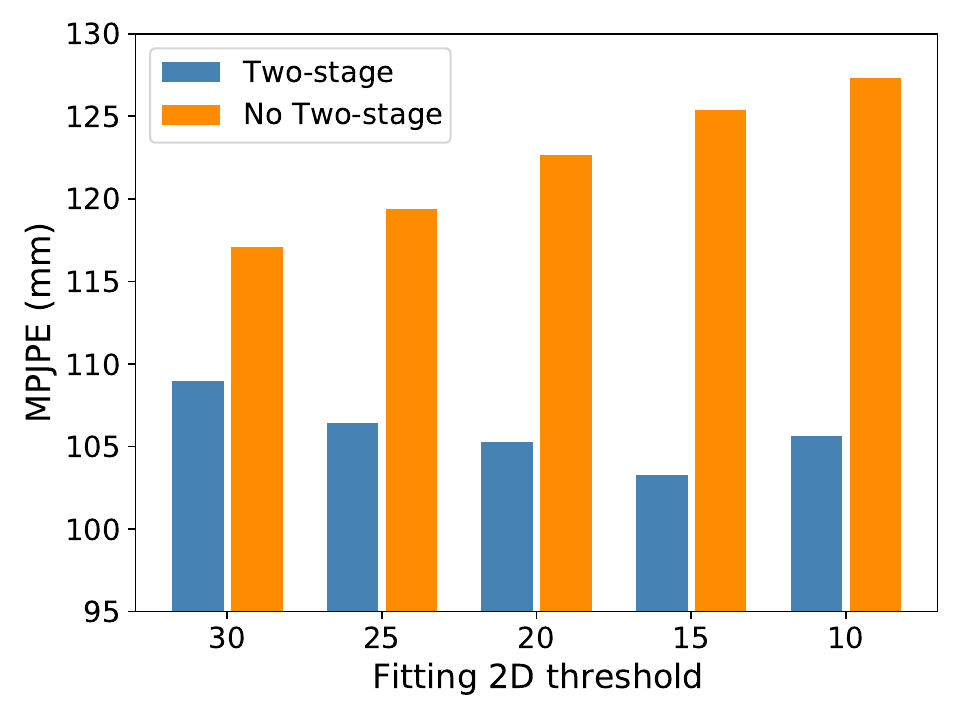}
	\caption{The two-stage optimization benefits from 2D projection while minimizing introduced errors.}
	\label{fig:two_stage}
    \vspace{-0.15in}
\end{figure}

\subsection{Ablation Studies}\label{sec:ablation}
Table~\ref{tab:pipelines_n_ablation} (a) shows that adaptive aggregation, local augmentation, and two-stage optimization all bring improvements.

\noindent \textbf{Sample selection in adaptive aggregation.}
Intuitively, the samples used 
should have low error and be representative of the current video. 
We find low-error samples by selecting those with high sampling weight;
however, exclusively relying on sampling weight will limit pose diversity.
Therefore, we separate samples into confident and non-confident subsets and adopt a balanced selection.
To maintain pose distribution, we add pose clustering finally.
As shown in Table~\ref{tab:pipelines_n_ablation} (b), uniform sampling performs worse than weight-based sampling, but the best results are obtained by combining balanced selection and pose clustering.

\noindent\textbf{Other components of adaptive aggregation.}
To leverage the information based on sample difficulty,
we used weak and strong pseudo-labeling adaptively (see Eq.~\ref{Eq:adaptive_pl}).
Table~\ref{tab:pipelines_n_ablation} (b) shows that using either only weak or only strong pseudo-labeling does not yield better results compared to a strategy where confident samples are subjected to strong pseudo-labeling and non-confident samples to weak pseudo-labeling.

\begin{figure*}[!htp]
	\centering
    \begin{minipage}[t]	{0.95\linewidth} \includegraphics[width=\linewidth]{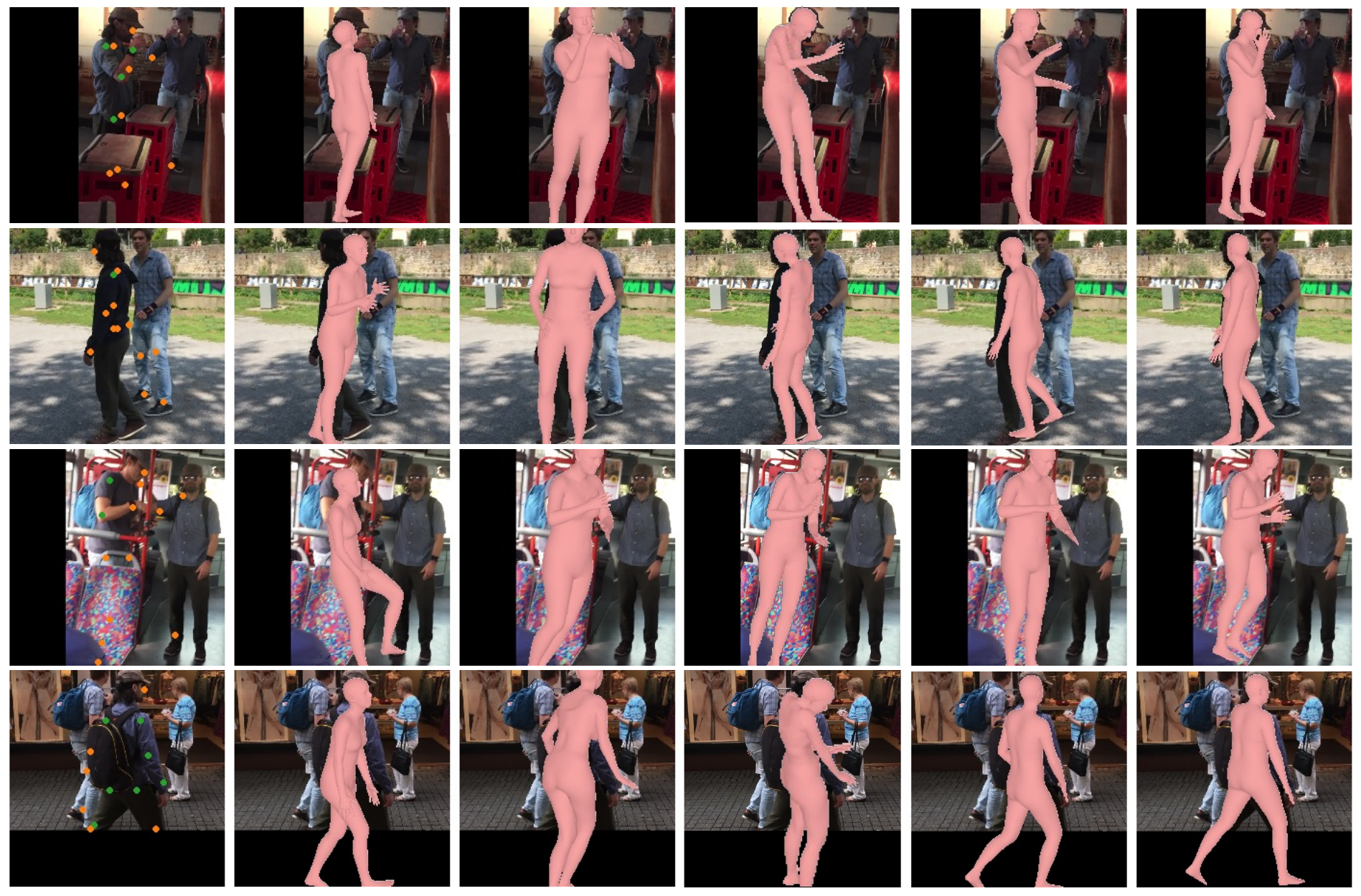}
    \end{minipage}\\
    \begin{minipage}[t]{\linewidth}
    \begin{minipage}[t]{0.21\linewidth} 
    \centering \small(a) {Estimated 2D over Image}
    \end{minipage}%
    \begin{minipage}[t] {0.13\linewidth} 
    \centering \small(b) {EFT}
    \end{minipage}%
    \begin{minipage}[t]
    {0.17\linewidth} \centering\small (c) {DynaBOA}
        \end{minipage}%
    \begin{minipage}[t]
    {0.13\linewidth} 
    \centering \small(d) {CycleAdapt*}
        \end{minipage}%
    \begin{minipage}[t]
    {0.18\linewidth} 
    \centering \small(e) {Ours}
        \end{minipage}%
    \begin{minipage}[t]
    {0.12\linewidth} 
    \centering \small(f) {Ground Truth}
    \end{minipage}%
    \end{minipage}
	\caption{Qualitative results on a 3DPW test sample comparing our method with EFT~\cite{eft}, DynaBOA~\cite{dynaboa} and CycleAdapt*~\cite{cycleadapt}. We present the estimated 2D results (a) with confident keypoints colored green and non-confident keypoints colored orange. Our method shows better results in challenging cases involving truncation and complex backgrounds. *Modified CycleAdapt source code to accommodate the online setting.}
	\label{fig:visualization}
    \vspace{-0.1in}
\end{figure*}

\noindent\textbf{Balance 2D projection and minimizing incorrect updates.}
In our two-stage optimization, the second stage is stopped based on a 2D projection threshold, as defined in Eq.~\ref{Eq:2D}.
Fig.~\ref{fig:two_stage} shows that, in the two-stage setting, reducing the 2D projection threshold (from 30 to 15) to emphasize the projection guidance leads to improved results, whereas an overly strict threshold of 10 leads to the model incorporating excessive noise.
However, without the two-stage optimization, wrongly estimated 2D poses leads to overall impaired performance. Our two-stage optimization effectively prevents this by maintaining 2D projection benefits and reducing the impact of errors through separation after the first stage.

\subsection{Qualitative Results}~\label{sec:qualitative}
Fig.~\ref{fig:visualization} visualizes the results of our method compared to others on challenging samples with low-confidence 2D inputs. Our approach demonstrates superior performance by producing more accurate 3D human mesh reconstructions, even in cases with insufficient illumination or truncation. This highlights the robustness of our method in handling difficult scenarios where other methods often fail to provide reliable predictions.

\section{Conclusion and Limitations}

Adapting the 3D pose model to unseen streaming data with estimated 2D poses is a challenging and practical task. 
To better solve this problem, we identify the  challenge of limiting estimation errors while preserving pose information when using estimated 2D poses.
Based on that, we propose adaptive aggregation, a two-stage optimization, and local augmentation for handling varying levels of estimated pose error.
First, we propose adaptive aggregation to re-initialize the 3D model and leverage useful information from the historical data. The influence of wrong updates in each sample adaptation is further minimized through a two-stage optimization. For the non-confident 2D labels that can hardly refine their 3D predictions, we propose local augmentation, which performs augmentation on adjacent confident samples to help the adaptation of the current non-confident sample. Extensive experiments demonstrate the effectiveness of each component and its significant improvement over state-of-the-art methods over two large-scale benchmarks.

However, our method still depends on the quality of the estimated 2D poses.  A better solution would be to enhance the 2D data quality during model adaptation as well. 
Additionally, we use temporally adjacent samples to assist non-confident samples; this strategy could be improved by incorporating richer temporal information from multiple adjacent samples. Addressing these limitations will be future work.

{
    \small    \bibliographystyle{ieeenat_fullname}    \bibliography{main.bib}
}

\end{document}